\documentclass[10pt,twocolumn,letterpaper]{article}
\PassOptionsToPackage{dvipsnames}{xcolor}

\usepackage{afterpage}
\usepackage{multirow}
 \usepackage[pagenumbers]{iccv} 
\usepackage[section]{placeins}
\definecolor{darkred}{RGB}{192,0,0}
\definecolor{darkblue}{RGB}{0,112,192}
\definecolor{iccvblue}{rgb}{0.21,0.49,0.74}
\usepackage{multirow}
\usepackage[pagebackref,breaklinks,colorlinks,allcolors=iccvblue]{hyperref}
\usepackage{colortbl}
\def\paperID{7462} 
\def\confName{ICCV}
\def\confYear{2025}

\title{DALIP: Distribution Alignment-based Language-Image Pre-Training for Domain-Specific Data}

\author{\fontsize{10pt}{11pt}\selectfont Junjie Wu\textsuperscript{1}\quad 
Jiangtao Xie\textsuperscript{2}\quad
Zhaolin Zhang\textsuperscript{1}\quad
Qilong Wang\textsuperscript{1,\thanks{Corresponding author. E-mail: \{wjj\_, qlwang\}@tju.edu.cn. \newline
\hspace*{1.5em} Project page: \href{https://github.com/XavierHeart/DALIP}{https://github.com/XavierHeart/DALIP}.}} 
Qinghua Hu\textsuperscript{1}\quad
Peihua Li\textsuperscript{2}\quad
Sen Xu\textsuperscript{3}
\\
\fontsize{10pt}{11pt}\selectfont 
\textsuperscript{1}Tianjin University\quad\textsuperscript{2}Dalian University of Technology\quad\textsuperscript{3}Yancheng Institute of Technology\\
}
\begin{document}
\maketitle
\begin{abstract}
Recently, Contrastive Language-Image Pre-training (CLIP) has shown promising performance in domain-specific data (\eg, biology), and has attracted increasing research attention. Existing works generally focus on collecting extensive domain-specific data and directly tuning the original CLIP models. Intuitively, such a paradigm takes no full consideration of the characteristics lying in domain-specific data (\eg,  fine-grained nature of biological data) and so limits model capability, while mostly losing the original ability of CLIP in the general domain. In this paper, we propose a Distribution Alignment-based Language-Image Pre-Training (DALIP) method for biological data. Specifically, DALIP optimizes CLIP models by matching the similarity between feature distribution of image-text pairs instead of the original [cls] token, which can capture rich yet effective information inherent in image-text pairs as powerful representations, and so better cope with fine-grained nature of biological data. Particularly, our DALIP efficiently approximates feature distribution via its first- and second-order statistics, while presenting a Multi-head Brownian Distance Covariance (MBDC) module to acquire second-order statistics of token features efficiently. Furthermore, we collect a new dataset for plant domain (\ie, specific data in biological domain) comprising 10M plant data with 3M general-domain data (namely PlantMix-13M) according to data mixing laws. Extensive experiments show that DALIP clearly outperforms existing CLIP counterparts in biological domain, while well generalizing to remote sensing and medical imaging domains. Besides, our PlantMix-13M dataset further boosts performance of DALIP in plant domain, while preserving model ability in general domain. 
\end{abstract}    
\vspace{-1.0cm}
\section{Introduction}
\label{sec:intro}
\begin{figure}[ht]
\centering
\subfloat[ ]{\includegraphics[width=0.9\linewidth]{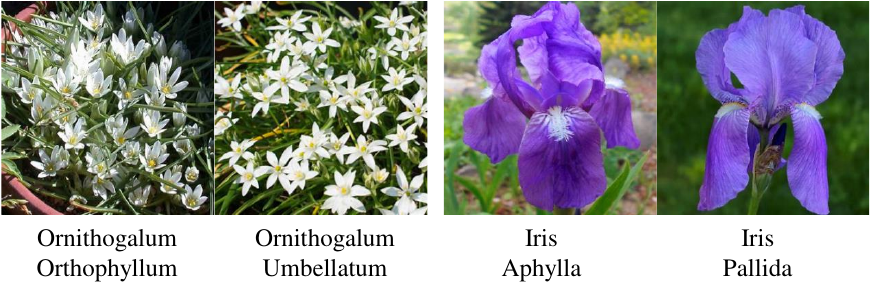}}
\newline
\subfloat[]{\includegraphics[width=0.93\linewidth]{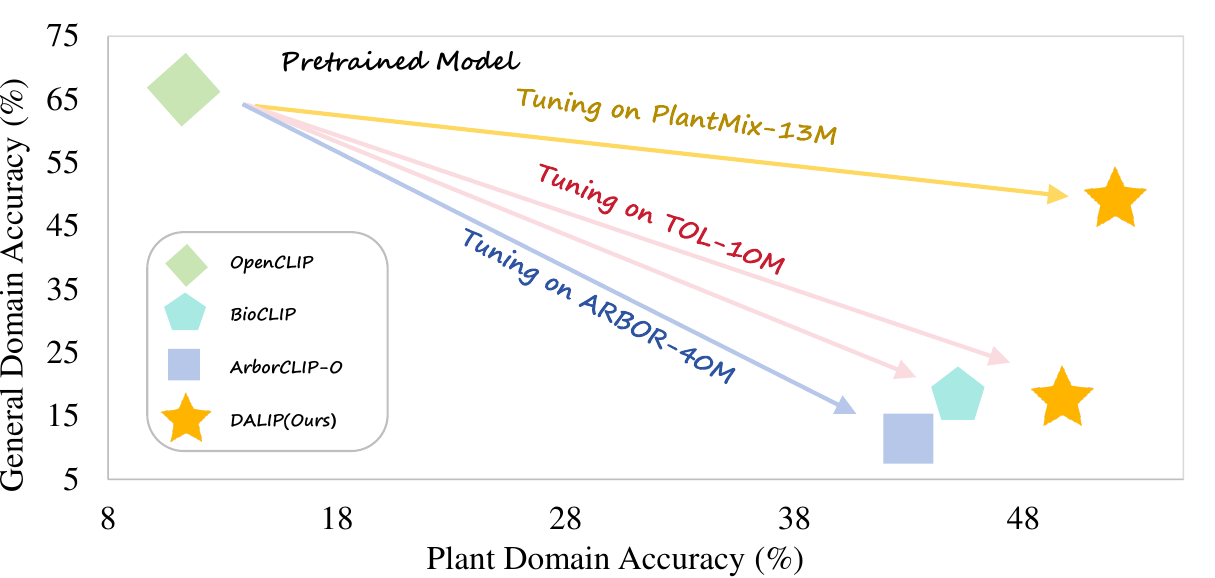}}
\vspace{-0.3cm}
\caption{(a) Example images in plant domain. (b) Comparison of different CLIP models on general (ImageNet-1K) and specific (average on five plant tasks) domains, including OpenCLIP~\cite{radford2021learning}, BioCLIP~\cite{stevens2024bioclip}, ArborCLIP-O~\cite{yang2024arboretum} and our DALIP, where DALIP achieves promising results in both plant and general domains.} 
\label{fig:intro}
\end{figure}
Recently, accompanied by the success of Contrastive Language Image Pre-training (CLIP)~\cite{radford2021learning}, domain-specific
CLIP models have achieved remarkable progress and attracted increasing research attention, such as biological~\cite{stevens2024bioclip,yang2024arboretum}, remote sensing~\cite{liu2024remoteclip,DBLP:journals/tgrs/ZhangZGY24}, and medical domains~\cite{eslami2021does,wang2022medclip,zhang2022contrastive,zhang2023biomedclip}. Particularly, they show the CLIP models trained in general domain~\cite{radford2021learning,DBLP:conf/iclr/0001XT0HS0GZF24} have underperformed in a certain domain-specific task, and make several efforts to enhance the ability of CLIP models in domain-specific applications. To this end, existing works generally focus on constructing large-scale training datasets~\cite{stevens2024bioclip,yang2024arboretum,DBLP:journals/tgrs/ZhangZGY24,zhang2023biomedclip} involving millions of domain-specific image-text pairs, and then employ them to directly tune the CLIP models~\cite{radford2021learning,DBLP:conf/iclr/0001XT0HS0GZF24}.

Although promising performance has been achieved, there still exist two challenges in domain-specific CLIP models. First, domain-specific data usually possess distinct some characteristics. By taking biological images (\eg, plant) as an example, as illustrated in Fig.~\ref{fig:intro} (a), they are fine-grained and large-species~\cite{stevens2024bioclip,van2021inat}. Therefore, compared with images in general domain, biological ones generally require more powerful and discriminative representations to capture subtle differences for accurate identification~\cite{DBLP:journals/pami/WeiSAWPTYB22}. However, data characteristics are barely considered in previous domain-specific CLIP models carefully. Second, existing domain-specific CLIP models mostly lose the original ability in the general domain, which poses the issue of catastrophic forgetting~\cite{DBLP:conf/acl/ZhengQ024,cclip} and limits application scenarios of domain-specific CLIP. Particularly, CLIP as a fundamental part of popular multimodal large language models~\cite{qwen2,DBLP:journals/corr/abs-2412-05271} is desirable to strong ability in both specific and general domains~\cite{DBLP:journals/corr/abs-2309-10313}. As shown in Fig.~\ref{fig:intro} (b), BioCLIP~\cite{stevens2024bioclip} clearly improves CLIP on biological tasks (\ie, five plant tasks in Meta-Album~\cite{ullah2022meta}), while the performance of zero-shot ImageNet classification decreases from 67.0\% to 18.6\%, suffering from severe performance loss.

In this paper, we propose a Distribution Alignment-based Language-Image Pre-Training method, namely DALIP, to address the above challenges specific to biological data. Specifically, the core idea of our DALIP is to model feature distribution of both image and text as powerful representations instead of a single [\textit{cls}] token in the original CLIP, and then optimize CLIP models by matching the similarity between feature distribution of image-text pairs. For computational efficiency, we approximate feature distribution by using the first two terms of its characteristic function, corresponding to first- and second-order statistics of token features. Meanwhile, DALIP presents a Multi-head Brownian Distance Covariance (MBDC) module to acquire second-order statistics of token features in an efficient way, which computes Brownian Distance Covariance (\ie, second-order statistics)~\cite{DBLP:conf/cvpr/XieLLWL22} within each group of token features following by a two-layer fully-connected layer for information interaction between groups. As such, our DALIP can efficiently capture effective information inherent in image-text pairs, which results in more discriminative representations and better copes with the fine-grained nature of biological data. Fig.~\ref{fig:intro} (b) shows DALIP clearly improves previous works~\cite{stevens2024bioclip,yang2024arboretum} with same or less training data. 

To alleviate the issue of catastrophic forgetting, we collect a new dataset (namely PlantMix-13M) for plant domain (\ie, specific data in biological domain) according to data mixing laws~\cite{DBLP:journals/corr/abs-2403-16952}. Specifically, our PlantMix-13M comprises 10M plant data from iNaturalist~\cite{van2021inat,van2018inaturalist} and DataComp1B~\cite{gadre2024datacomp} with 3M general-domain data from DataComp1B~\cite{gadre2024datacomp}. To generate text descriptions for plant images, we prompt a fine-tuned Qwen2-VL-7B~\cite{Qwen2VL} with GPT-4o by using latin and common names to give language description on key visual features in the plant images. Besides, we also combine the text descriptions of \textit{"a photo of [latin name], commonly known as [common name]"} as suggested in~\cite{stevens2024bioclip}. As shown in Fig.~\ref{fig:intro} (b), our PlantMix-13M can help DALIP models achieve better performance trade-offs between plant and general domains. To evaluate our methods, we conduct experiments on 15 downstream tasks in both biological (\eg, Meta-Album~\cite{ullah2022meta}) and general (\eg, ImageNet-1K~\cite{DBLP:conf/cvpr/DengDSLL009}) domains, where various CLIP models (\eg, OpenCLIP~\cite{radford2021learning} and SigLIP~\cite{zhai2023sigmoid}) are trained on TREEOFLIFE-10M~\cite{stevens2024bioclip} and our PlantMix-13M. The contributions of our work are concluded as follows.

\noindent $\spadesuit$ In this paper, we propose a Distribution Alignment-based Language-Image Pre-Training (DALIP) method to improve the performance of CLIP models in biological domain, whose core idea is to generate powerful representations for image-text pairs for effectively optimizing CLIP models. Particularly, an MBDC module is presented to efficiently acquire second-order statistics of token features.

\noindent $\spadesuit$ Besides, we construct a PlantMix-13M dataset for plant domain, allowing our DALIP models to achieve better performance trade-offs between plant and general domains. To our best knowledge, our work makes the first attempt to preserve the ability of specific-domain CLIP models in the general domain under the practical large-scale setting.

\noindent $\spadesuit$ Extensive experiments are conducted on various downstream tasks in both biological and general domains, and the results show our DALIP clearly outperforms existing CLIP counterparts in biology domain. Besides, our PlantMix-13M dataset further boosts the performance of DALIP in plant domain, while preserving the ability of the CLIP models in the general domain. Additionally, experiments on remote sensing and medical imaging domains further verify the generalization of our DALIP. 

\section{Related Work}
\label{sec:related}
\textbf{Domain-specific CLIP Models.} The advanced performance of Contrastive Language-Image Pre-training~\cite{radford2021learning}  (CLIP) accelerates several domain-specific applications. Specifically, CLIP is first adapted in the medical domain, and several domain-specific CLIP models are proposed, such as PubMedCLIP~\cite{eslami2021does}, MedCLIP~\cite{wang2022medclip}, ConVIRT~\cite{zhang2022contrastive} and BioMedCLIP~\cite{zhang2023biomedclip}. Among them, PubMedCLIP and MedCLIP tune CLIP models on MIMIC-CXR dataset~\cite{DBLP:journals/corr/abs-1901-07042} of about 200K image-text pairs, where MGCA~\cite{DBLP:conf/nips/WangZWVY22} trains a CLIP-style model. BioMedCLIP trains CLIP models on a newly built PMC-15M dataset of 15 million biomedical image-text pairs. Subsequently, GeoRSCLIP~\cite{DBLP:journals/tgrs/ZhangZGY24} and RemoteCLIP~\cite{liu2024remoteclip} are proposed in remote sensing domain, which tunes CLIP models on RS5M dataset of 5 million image-text pairs and a dataset of 828,725 image-text pairs, respectively. Recently, BioCLIP~\cite{stevens2024bioclip} and ArborCLIP~\cite{yang2024arboretum} achieve promising performance in the biological domain, where CLIP models are tuned on newly-built TREEOFLIFE-10M and ARBORETUM-40M datasets, respectively. Different from the aforementioned works, we propose a DALIP method for biological domain by carefully considering their data characteristics while collecting a PlantMix-13M dataset to balance performance between plant and general domains.

\noindent\textbf{CLIP Variants.} The original  CLIP~\cite{radford2021learning} and ALIGN~\cite{jia2021scaling} are trained on large-scale image-text datasets~\cite{DBLP:journals/corr/abs-2111-02114} by using softmax contrastive learning~\cite{zhang2022contrastive,oord2018representation,chen2020simple,khosla2020supervised}. Going beyond them, several variants are proposed to enhance the CLIP models from the perspectives of data processing~\cite{yang2023alip,DBLP:conf/iclr/0001XT0HS0GZF24}, efficient training~\cite{zhai2022lit,li2023blip,moayeri2023text} and modification of training objectives~\cite{zhai2023sigmoid,desai2023hyperbolic,gao2024embracing,chun2024probabilistic}. Specifically, MetaCLIP~\cite{DBLP:conf/iclr/0001XT0HS0GZF24} is introduced to utilize metadata expansion and create a substantial CommonCrawl dataset of 400 million image-text pairs. SigLIP~\cite{zhai2023sigmoid} improves training efficiency by introducing a pairwise sigmoid loss. For fine-grained visual-language alignment, FLAIR~\cite{DBLP:journals/corr/abs-2412-03561} enhances region-phrase correspondence through cross-attention between local image regions and text tokens. MGCA~\cite{DBLP:conf/nips/WangZWVY22} leverages masked modeling and cross-modal augmentation to capture fine-grained features for medical tasks. EAU~\cite{gao2024embracing} and ProLIP~\cite{chun2024probabilistic} introduce the idea of probability distribution matching into CLIP, which is closely related to our DALIP. However, both of them compute variance of distribution by using a learnable vector (not exact variance of token features), which takes no full advantage of rich statistics lying in token features. Different from them, our DALIP introduces an MBDC module to capture precise second-order statistics of token features efficiently. As  shown in Table \ref{tab:Plantmix} and Table \ref{tab:cross_domain}, DALIP clearly outperforms the aforementioned methods.

\section{DALIP Method}
\label{sec:method}
In this section, we first revisit the original CLIP, and then introduce distribution alignment for CLIP via first- and second-order statistics approximation. Finally, we present a Multi-head Brownian Distance Covariance (MBDC) to acquire second-order statistics of token features efficiently.

\begin{figure*}[!ht]
\vspace{-0.4cm}
\centering
\includegraphics[width=0.95\linewidth]{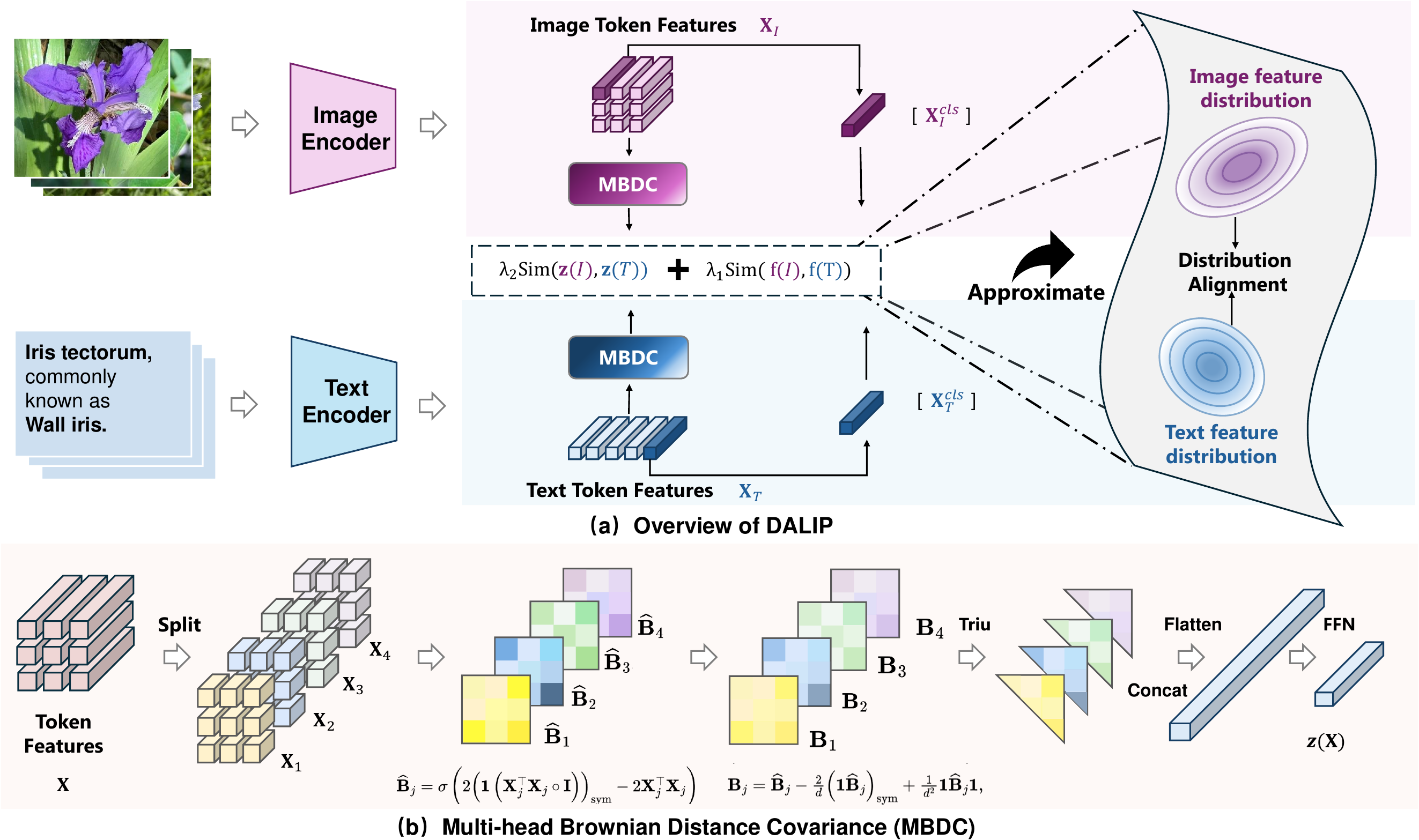}
\caption{(a) Overview of our Distribution Alignment-based Language-Image Pre-Training (DALIP) method for biological data. Specifically, DALIP optimizes CLIP models by matching the similarity between feature distribution of image-text pairs, which are efficiently approximated by first- and second-order statistics of token features. Particularly, (b) a Multi-head Brownian Distance Covariance (MBDC) module is presented to efficiently acquire second-order statistics of token features, whose details can be found in Sec.~\ref{sec:MDBC}.}
\label{fig:Framework-all} 
\end{figure*}

\subsection{Revisit CLIP}

Given a mini-batch training data $S=\left\{\left(I^i, T^i\right)\right\}_{i=1}^N$ of $N$ image-text pairs, the classical CLIP~\cite{radford2021learning} trains the models by using a multi-modal InfoNCE loss~\cite{DBLP:journals/corr/abs-1807-03748} to align $N$ image-text pairs, \ie, 
\begin{equation}
\label{equ-infornce}
\begin{aligned}
  -\sum_{i=1}^N \left[ 
    \log \frac{e^{f(I^i)^{\top} f(T^i) / \tau}}{\sum_j e^{f(I^i)^{\top} f(T^j) / \tau}} +\log \frac{e^{f(I^i)^{\top} f(T^i) / \tau}}{\sum_j e^{f(I^j)^{\top} f(T^i) / \tau}} \right],
\end{aligned}
\end{equation}
where  $f(I^i) \in \mathbb{R}^{1 \times d}$ and $f(T^i) \in \mathbb{R}^{N \times d}$ respectively indicate feature embeddings of image and text inputs, and  $\tau$ is the temperature parameter. Let $\mathbf{X}_{I^i}\in \mathbb{R}^{M \times d}$ be word token features~\cite{dosovitskiy2020image} or convolution features~\cite{liu2022convnet} in the last layer of image encoder, $f(I^i)$ is usually generated by [\textit{cls}] token~\cite{dosovitskiy2020image} (\ie, $f(I^i)=\mathbf{x}_{I^i}^{cls}$) or average pooling of convolution features~\cite{liu2022convnet}) ($f(I^i)=\sum_{k=1}^{M}\mathbf{X}_{I^i}^{k}$). For text embeddings, [\textit{cls}] token~\cite{radford2019language} is used, \ie, $f(T^i)=\mathbf{x}_{T^i}^{cls}$. Intuitively, [\textit{cls}] token $\mathbf{x}_{\cdot}^{cls}$ can be regarded as a weighted average of word tokens ($\mathbf{x}_{\cdot}^{cls}=\mathbf{w}\mathbf{X}_{\cdot}$), where $\mathbf{w}$ are self-attention weights. Therefore, feature embeddings of image-text pairs in Eqn.~(\ref{equ-infornce}) only exploit first-order statistics of image and text features, while representation ability of feature embeddings potentially impacts training CLIP models. 

\subsection{Distribution Alignment for CLIP}
By considering the fine-grained nature in both visual cues and text descriptions of biological data, first-order statistics of image and text features hardly provide discriminative representations to capture subtle differences between different classes~\cite{DBLP:journals/pami/WeiSAWPTYB22}. Therefore, this paper proposes a Distribution Alignment-based Language-Image Pre-Training (DALIP) method for biological data, which computes feature distributions of image-text pairs as powerful representations to optimize CLIP objective, \ie, 
\begin{equation}   
\small
\label{equ-infornce1}
  -\sum_{i=1}^N \left[ 
    \log \frac{e^{\text{sim}(G(\mathbf{X}_I^{i}), G(\mathbf{X}_T^{i}))}}{\sum_j e^{\text{sim}(G(\mathbf{X}_I^{i}), G(\mathbf{X}_T^{j}))}} +\log \frac{e^{\text{sim}(G(\mathbf{X}_I^{i}), G(\mathbf{X}_T^{i}))}}{\sum_j e^{\text{sim}(G(\mathbf{X}_I^{j}), G(\mathbf{X}_T^{i}))}} \right],
\end{equation}
where $G(\mathbf{X}_I^{i})$ and $G(\mathbf{X}_T^{i})$ respectively indicate distributions of features $\mathbf{X}_I^{i}$ and $\mathbf{X}_T^{i}$, while sim means similarity between feature  distributions. Compared to the [\textit{cls}] token, feature distribution can portray characteristics of data more comprehensively and explore rich yet effective information lying in token features for coping with biological data.

For efficiently matching feature distributions $G(\mathbf{X}_I^{i})$ and $G(\mathbf{X}_T^{i})$, we first represent probability density function $G(\mathbf{X}^{i})$ (we omit the subscript for simplicity) by characteristic function $\psi_{\mathbf{X}^{i}}(m)$ with Taylor series~\cite{bishop2006pattern} as
 \begin{equation}
\psi_{\mathbf{X}^{i}}(m) = 1 + {\alpha}_1^{i}\mathbf{Q}_1^{i} + {\alpha}_2^{i}{\mathbf{Q}}_2^{i} + \cdots = 1 + \sum_{k=1}^{\infty}{\alpha}_k^i\mathbf{Q}_k^{i},
\label{formula 3}
\end{equation}
where $m \in \mathbb{R}$ is the argument of the characteristic function. $\mathbf{Q}_k^{i}$ and ${\alpha}_k^i$ indicate $k$-th order statistics of features $\mathbf{X}^{i}$ and its weight. Furthermore, as suggested in ~\cite{DBLP:conf/iccv/GaoWLZHZ23}, we compute first- and second-order statistics of features $\mathbf{X}^{i}$ to approximate the characteristic function $\psi_{\mathbf{X}^{i}}(m)$ (\textit{w.r.t} $G(\mathbf{X}^{i})$) for computational efficiency, \ie,
\begin{equation}
G(\mathbf{X}^{i}) \approx {\alpha}_1^{i}\mathbf{Q}_1^{i} + {\alpha}_2^{i}{\mathbf{Q}}_2^{i}. 
\label{formula 4}
\end{equation}

Therefore, the objective function of our DALIP in Eqn.~(\ref{equ-infornce1}) can be rewritten as
\begin{equation}
\scriptsize
\label{equ-mid}
\begin{aligned}
&-\sum_{i=1}^N \Bigg[ 
\textcolor{darkblue}{\lambda_1 \log \frac{e^{\mathbf{Q}_1(I^i)^{\top} \mathbf{Q}_1(T^i) / \tau}}{\sum_j e^{\mathbf{Q}_1(I^i)^{\top} \mathbf{Q}_1(T^j) / \tau}} 
+ \lambda_1 \log \frac{e^{\mathbf{Q}_1(I^i)^{\top} \mathbf{Q}_1(T^i) / \tau}}{\sum_j e^{\mathbf{Q}_1(I^j)^{\top} \mathbf{Q}_1(T^i)/ \tau}} } \\
&+ \textcolor{darkred}{\lambda_2 \log \frac{e^{\text{sim}(\mathbf{Q}_2(I^i), \mathbf{Q}_2(T^i))}}{\sum_j e^{\text{sim}(\mathbf{Q}_2(I^i), \mathbf{Q}_2(T^j))}} 
+ \lambda_2 \log \frac{e^{\text{sim}(\mathbf{Q}_2(I^i), \mathbf{Q}_2(T^i))}}{\sum_j e^{\text{sim}(\mathbf{Q}_2(I^j), \mathbf{Q}_2(T^i))}} } \Bigg],
\end{aligned}
\end{equation}
where the \textcolor{darkblue}{blue} and \textcolor{darkred}{red} terms are corresponding to the objective functions based on \textcolor{darkblue}{first-} and \textcolor{darkred}{second-order} statistics, respectively. $\lambda_{1}$ and $\lambda_{2}$ are trade-off parameters. The first-order statistics $\mathbf{Q}_1(I^i)$ and $\mathbf{Q}_1(T^i)$ for image and text features can be computed as [\textit{cls}] token or average pooling of word tokens, \ie, $f(I^i)$ and $f(T^i)$ in Eqn.~(\ref{equ-infornce}). In general, second-order statistics $\mathbf{Q}_2(I^i)$ ($\mathbf{Q}_2(T^i)$) can be achieved by outer product of word token features. 

\begin{figure*}[htb]
\centering
\vspace{-0.4cm}
\includegraphics[width=0.95\textwidth]{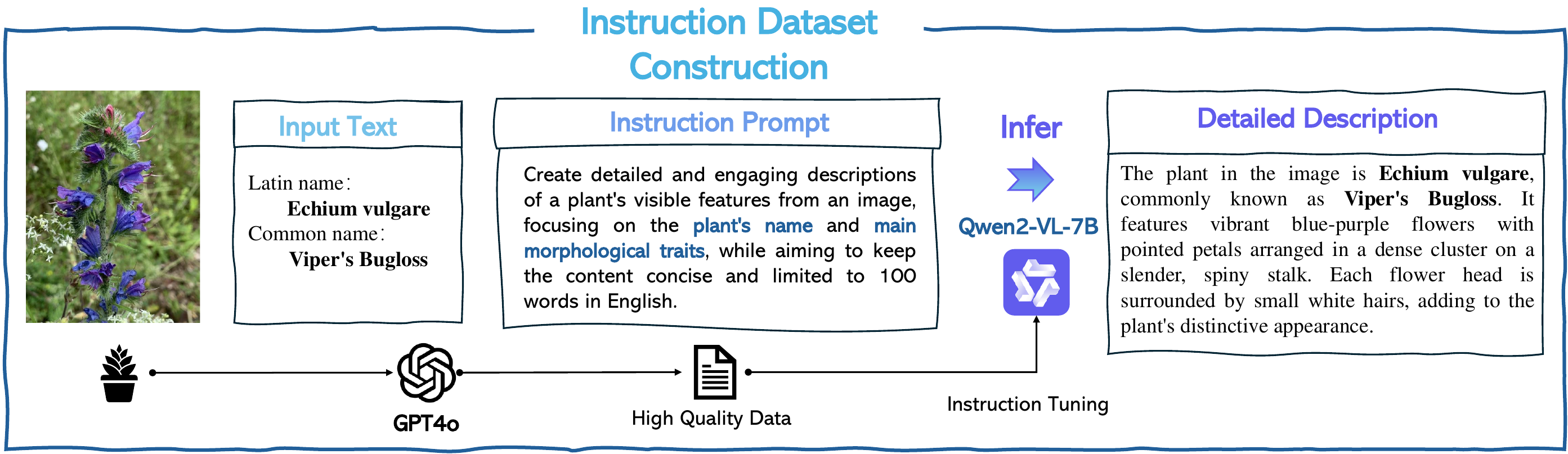}
\vspace{-0.2cm}
  \caption{Example illustration of generating detailed plant descriptions by prompting Qwen2-VL-7B~\cite{Qwen2VL}, where Latin and Common names, images, and tailored instruction prompts are used as inputs.}
  \label{fig:qwen}
  \vspace{-0.4cm}
\end{figure*}

\subsection{MDBC for Efficient Second-order Statistics}\label{sec:MDBC}
 By taking word token features of $i$-th image ($\mathbf{X}_{I^{i}}\in \mathbb{R}^{M\times d}$) as example, second-order statistics $\mathbf{Q}_2(I^i)$ achieved by the outer product of $\mathbf{X}_{I^{i}}$ generally suffer from two limitations. First, the outer product (also namely covariance) focuses on modeling linear correlation among different feature channels, also observed in~\cite{engin2018deepkspd}, limiting to capture nonlinear complex correlation and leading to imperfect second-order representations~\cite{DBLP:journals/ijcv/ZhangWZL21}. Second, $(\mathbf{X}_{I^{i}})^{\top}(\mathbf{X}_{I^{i}})$ results in a $d^2$-dimensional representation. For high-dimensional features (\eg, $d=768$ in \cite{radford2021learning}), such large-size representations will bring heavy computational cost. To handle above issues, we present a Multi-head Brownian Distance Covariance (MBDC) to acquire the second-order representations effectively and efficiently. As shown in Figure~\ref{fig:Framework-all} (b), MBDC first splits token features $\mathbf{X}$ (we omit the subscript for briefness) into $h$ heads along feature dimension: 
\begin{equation}
 [\mathbf{X}_1; \mathbf{X}_2; \cdots; \mathbf{X}_h] = {\rm SP}\big(\mathbf{X}\big),
\label{formula 7}
\end{equation}
where $\rm SP$ indicates a splitting operation, and $\mathbf{X}_j\in \mathbb{R}^{M\times (d/h)}$ are split features. Then, MBDC computes Brownian distance covariance~\cite{szekely2009brownian,DBLP:conf/cvpr/XieLLWL22} for each split feature as
\begin{equation}
\label{equ:bdc}
\begin{split}
\widehat{\mathbf{B}}_j &= \sigma\left(2\left(\mathbf{1}\left(\mathbf{X}_j^{\top} \mathbf{X}_j \circ \mathbf{I}\right)\right)_{\text{sym}} - 2 \mathbf{X}_j^{\top} \mathbf{X}_j\right), \\
\mathbf{B}_j &= \widehat{\mathbf{B}}_j - \frac{2}{d}(\mathbf{1} \widehat{\mathbf{B}}_j)_{\text{sym}} + \frac{1}{d^2} \mathbf{1} \widehat{\mathbf{B}}_j \mathbf{1}.
\end{split}
\end{equation}

where $\circ$ indicates dot product operation. $\mathbf{1}$ and $\mathbf{I}$ are an all-one vector and an identity matrix, respectively. sym means the matrix symmetry operation, and $\sigma$ is an element-wise square-root normalization. Particularly, $\mathbf{B}_j$ leads to a $(d/h)^2$-dimensional second-order representation, which can model a more complex correlation than the original covariance matrix. However, $\mathbf{B}_j$ neglects feature correlations among different heads, and so our MBDC introduces a layer normalization (LN) and a two-layer FFN to perform information interaction for multi-head representations:
\begin{equation}
\mathbf{z}=\text{FFN}_{\mathbf{W}_1,\mathbf{W}_2}\left(\rm LN\left(\rm Concat[\Tilde{\mathbf{B}}_1; \Tilde{\mathbf{B}}_2; \cdots; \Tilde{\mathbf{B}}_h]\right)\right),
\label{formula 8}
\end{equation}
where $\Tilde{\mathbf{B}}_j$ indicates upper triangulation of $\mathbf{B}_j$, leading to a $l$-dimensional representation, where $l=((d/h)+1)(d/h)/2$. $\rm Concat$ is a concatenated operation. $\mathbf{W}_1\in \mathbb{R}^{l\times l}$ and $\mathbf{W}_2\in \mathbb{R}^{l\times \Tilde{d}}$ are the weights of FFN. Finally, our MBDC obtains a $\Tilde{d}$-dimensional second-order representation ($\Tilde{d}\ll d^2$). 

According to Eqn.~(\ref{formula 8}), we can calculate the similarity between second-order statistics (\ie, $\text{sim}(\mathbf{Q}_2(I^i), \mathbf{Q}_2(T^i)$) in Eqn.~(\ref{equ-mid}) by $\mathbf{z}(I^i)^{\top} \mathbf{z}(T^i)$. Finally, the total objective function of our DALIP in Eqn.~(\ref{equ-infornce1}) can be rewritten as

\begin{equation}
\small
\begin{aligned}
\label{equ-loss}
&-\sum_{i=1}^N \left[
  \textcolor{darkblue}{
    \lambda_1 \log \frac{e^{f(I^i)^{\top} f(T^i) / \tau}}{\sum_j e^{f(I^i)^{\top} f(T^j) / \tau}} +\lambda_1\log \frac{e^{f(I^i)^{\top} f(T^i) / \tau}}{\sum_j e^{f(I^j)^{\top} f(T^i) / \tau}} }\right. \\
    &\textcolor{darkred}{  + \left.  \lambda_2\log S \frac{e^{\mathbf{z}(I^i)^{\top} \mathbf{z}(T^i) / \tau}}{\sum_j e^{\mathbf{z}(I^i)^{\top} \mathbf{z}(T^j) / \tau}} +\lambda_2\log \frac{e^{\mathbf{z}(I^i)^{\top} \mathbf{z}(T^i) / \tau}}{\sum_j e^{\mathbf{z}(I^j)^{\top} \mathbf{z}(T^i) / \tau}}  \right].}
\end{aligned}
\end{equation}

Clearly, our DALIP trains the language-image models by aligning feature distributions, which are approximated by first- and second-order statistics of token features. Besides, DALIP (\ref{equ-loss}) can be flexibly optimized by the classical BP.
\section{PlantMix-13M Dataset}\label{sec:plantmix}
\begin{table}[t!]
\centering
\resizebox{0.475\textwidth}{!}{
\begin{tabular}{@{\extracolsep{\fill}}lccccc@{}}
\toprule
&Percentage & Text Details & Images & Classes & Source \\
\midrule
\multirow{3}{*}{\rotatebox[origin=c]{90}{Plants}} 
& 38.5\% & Brief Plant Latin and Common Names & 5M & 9318 & iNaturalist \\
\cmidrule{2-6}
& 30.8\% & Qwen2-VL-7B Detailed Plant Descriptions & 4M & 9554 & iNaturalist \\
\cmidrule{2-6}
& 7.6\% & Plant-specific Captions from Datacomp1B & 1M & - & Datacomp1B \\
\midrule
\rotatebox[origin=c]{90}{General}
& 23.1\% & General Datacomp1B Captions & 3M & - & Datacomp1B \\
\bottomrule
\end{tabular}
}
\vspace{-0.3cm}
\caption{Detailed composition of PlantMix-13M dataset.}
\label{table:dataset_composition}
\end{table}

\noindent \textbf{Data Collection.} Different from previous works~\cite{stevens2024bioclip,yang2024arboretum} mainly focus on ability of CLIP models in the specific domain, our models aim to preserve ability of CLIP models in the general domain while achieving promising performance in the specific domain. To this end, our paper constructs a PlantMix-13M dataset for plant-specific domain. Table~\ref{table:dataset_composition} summarizes the outline of our PlantMix-13M dataset. Specifically, it is composed of 10M plant images and 3M general images. For plant-specific data, 9M of them are collected from iNaturalist~\cite{van2018inaturalist}, and the remaining 1M images are collected from Datacomp1B~\cite{gadre2024datacomp}. Particularly, 1M plant-specific data from Datacomp1B can be a useful complement to iNaturalist, and enhance diversity and coverage of plant species. Since the images in iNaturalist are well annotated, we can provide a brief text description for each image in the format "a photo of [Latin name], commonly known as [common name]". 
To enhance text description quality, we develop a two-stage generation process. First, we manually refine 1,000 plant descriptions generated by GPT-4o to ensure botanical accuracy, aiming to create high-quality reference samples. They are further used to fine-tune Qwen2-VL-7B~\cite{Qwen2VL} for high-efficiency and low-cost generation, while enabling to generate comprehensive plant descriptions based on scientific and common names (Fig.~\ref{fig:qwen}). To satisfy the requirement of diverse yet comprehensive text descriptions~\cite{stevens2024bioclip}, we randomly select 4M images from iNaturalist, and annotate them with MLLM-generated comprehensive descriptions to increase text diversity. Additionally, 3M general-domain images from Datacomp1B are randomly sampled for data mixing, and mixing strategies are detailed as follows.

\begin{figure}[t!]
\centering
\vspace{-0.2cm}
\includegraphics[width=0.30\textwidth]{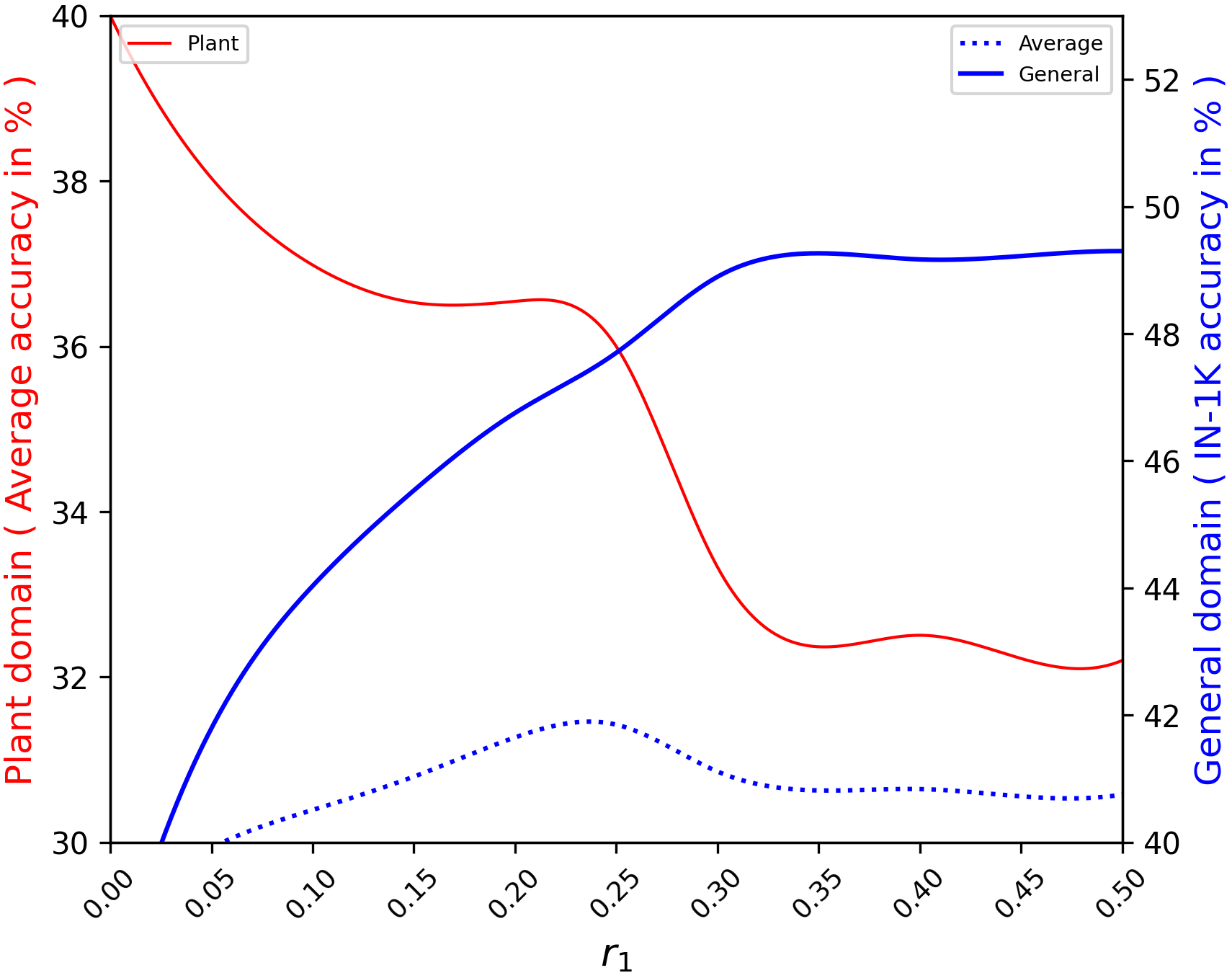}
  \caption{Results of various data mixing ratios in general (\ie, IN-1K) and plant domains (average on five tasks in Table~\ref{tab:Plantmix}).}
  \label{fig:datamix}
  \vspace{-0.5cm}
\end{figure}

\noindent \textbf{Data Mixing.}  
To efficiently determine the optimal data mixing ratio, our PlantMix-13M dataset is constructed based on a generalization-guided data mixing law, inspired by~\cite{DBLP:journals/corr/abs-2403-16952}. Particularly, we employ accuracies on downstream tasks as an indicator instead of training loss, and the relationship between the accuracies in different domains and data ratio $r_i$ can be calculated by
\begin{equation}
P_i(r_i) = \alpha_i + \beta_i \exp(\gamma_{ii}r_i),
\end{equation}
where $P_i$ represents the accuracy in domain $i$ with data ratio $r_i$, $\alpha_i$, $\beta_i$ and $\gamma_{ii}$ are fitting parameters. To obtain the fitting parameters $\alpha_i$, $\beta_i$ and $\gamma_{ii}$, we conduct experiments by fixing 1M plant data and varying proportions of general data. For accuracies in general and plant domains, we report zero-shot classification performance on ImageNet-1K and five plant-specific tasks. As observed in Fig~\ref{fig:datamix}, we can fit the data mixing laws for our case as

\begin{align}
P_1(r_1) &= 49.74 - 19.65 \exp(-9.46r_1), \nonumber \\
P_2(1-r_1) &= 89.9 - 71.6\exp(-0.36(1-r_1)),
\label{equ:acc}
\end{align}
where $r_1$ indicates data ratio in general domain. As shown in Eqn.~(\ref{equ:acc}), the best trade-off between $P_1(r_1)$ and $P_2(r_2)$ is achieved when $r_1$ is about 0.23. It suggests that the ratio of 10:3 for general and plant-specific data, leading to the construction of our PlantMix-13M with 10M plant-specific data and 3M general data. 


\section{Experiment}
In this section, we first describe the implementation details of our DALIP. Then, we compare with several state-of-the-art CLIP models by tuning on TOL~\cite{stevens2024bioclip} and our PlantMix datasets. Finally, ablation studies are conducted on our PlantMix datasets to assess the effect of key components.

\definecolor{c1}{HTML}{70f3ff}
\definecolor{c2}{HTML}{ffb3a7}
\begin{table*}[htb]
\centering
\small
\renewcommand{\arraystretch}{0.6}  
\begin{tabular}{lcccccccccccccc}
\toprule
\multirow{10}{*}{\textbf{Model}} & \multirow{10}{*}{\textbf{Dataset}} & \multicolumn{5}{c}{\textbf{Animals}} & \multicolumn{5}{c}{\textbf{Plants \& Fungi}}&\multirow{10}{*}{{\textbf{Mean}}}  &\multirow{10}{*}{ ($\Delta$)}  \\
\cmidrule(lr){3-7} \cmidrule(lr){8-12}
 && \rotatebox[origin=c]{90}{Birds525} & \rotatebox[origin=c]{90}{Plankton} & \rotatebox[origin=c]{90}{Insects} & \rotatebox[origin=c]{90}{Insects 2} & \rotatebox[origin=c]{90}{BCR} & \rotatebox[origin=c]{90}{PlantNet} & \rotatebox[origin=c]{90}{Fungi} & \rotatebox[origin=c]{90}{PlantVillage} & \rotatebox[origin=c]{90}{Med. Leaf} & \rotatebox[origin=c]{90}{PlantDoc} & \\
\midrule
\textcolor{gray}{Random Guessing} & --  & 0.2 &\textcolor{gray}{1.2} & \textcolor{gray}{1.0} & \textcolor{gray}{1.0 }&\textcolor{gray}{0.3}  &\textcolor{gray}{4.0}  & \textcolor{gray}{4.0 }& \textcolor{gray}{2.6} &\textcolor{gray}{4.0}  &\textcolor{gray}{3.7}  & \textcolor{gray}{2.4} \\
\midrule
CLIP~\cite{radford2021learning} & --&  49.9& 3.2 & 9.1 & 9.8 & 31.8 & 58.5 & 10.2 & 5.4 & 15.9 & 26.1 & 22.0& -- \\
OpenCLIP~\cite{DBLP:conf/cvpr/ChertiBWWIGSSJ23} & -- & 54.7 & 2.2 & 6.5 & 9.6 & 29.8 & 50.2 & 5.7 & 8.0 & 12.4 & 25.8 & 20.5& -- \\
\midrule
\multicolumn{14}{c}{Tuning on 1M dataset} \\
\midrule
BioCLIP~\cite{stevens2024bioclip} & TOL-1M& \textbf{65.1} & 3.5 & 30.6 & 17.3 & 30.9 & \textbf{86.3} & 32.8 & 19.9 & 18.7 & \textbf{24.5} & 33.0 & --\\
DALIP (Ours) & TOL-1M & 60.9 &\textbf{10.4} & \textbf{34.5} &\textbf{35.3}  & \textbf{32.1} & 82.5 & \textbf{41.8} &\textbf{22.5}  & \textbf{27.3} & 21.0 & \textbf{36.8}& + 3.8\\
\midrule
\multicolumn{14}{c}{Tuning on 10M dataset} \\
\midrule
BioCLIP~\cite{stevens2024bioclip} & TOL-10M & \textbf{72.1} & 6.1 & 34.8 & 20.4 & 30.2 & 91.4 & 40.7 & 24.4 & 38.6 & \textbf{28.4} & 38.7 & --\\
ArborCLIP-O~\cite{yang2024arboretum} & ARBOR-40M & 47.1 &9.1 & 41.3 & 40.5 & 22.9 & 64.5 & 40.1 & 19.6 & 38.0 & 20.5 & 34.4& - 4.3 \\
DALIP (Ours) & TOL-10M& 71.9 & \textbf{15.5} & \textbf{49.5} & \textbf{43.1} & \textbf{34.3} &\textbf{92.9}  & \textbf{56.5} & \textbf{25.9} & \textbf{46.3 }& 26.6 & \textbf{46.3} & + 7.6 \\
\bottomrule
\end{tabular}
\vspace{-0.2cm}
\caption{Performance comparison of various models on zero-shot classification tasks across animal and plant domains. BCR stands for the BioCLIP Rare species benchmark (n-classes=400).}
\label{tab:biological-domain}
\end{table*}

\definecolor{c1}{HTML}{70f3ff}
\begin{table*}
\centering
\small
\resizebox{\textwidth}{!}{
\begin{tabular}{lcccccccccccccccc}
\toprule
\multirow{10}{*}{\textbf{Model}} & \multirow{10}{*}{\textbf{Arch.}} & \multirow{10}{*}{\textbf{Dataset}} & \multicolumn{5}{c}{\textbf{General}}& \multirow{6}{*}{\textbf{\rotatebox[origin=c]{90}{General Mean}}}  & \multicolumn{5}{c}{\textbf{Plants \& Fungi}} &\multirow{7}{*}{\textbf{\rotatebox[origin=c]{90}{Plant Mean}}}& \multirow{10}{*}{\textbf{Mean}} \\
\cmidrule(lr){4-8} \cmidrule(lr){10-14}
 & & & \rotatebox[origin=c]{90}{ImageNet-1K} & \rotatebox[origin=c]{90}{Cifar-100} & \rotatebox[origin=c]{90}{Cars} & \rotatebox[origin=c]{90}{Pets} & \rotatebox[origin=c]{90}{Sun397} &&  \rotatebox[origin=c]{90}{PlantNet} & \rotatebox[origin=c]{90}{Fungi} & \rotatebox[origin=c]{90}{PlantVillage} & \rotatebox[origin=c]{90}{Med. Leaf} & \rotatebox[origin=c]{90}{PlantDoc} & \\
\midrule
CLIP~\cite{radford2021learning} & \multirow{2}{*}{ViT-B/16} & -- & 68.3 & 68.2 & 64.6 & 88.9 & 64.0& 70.8 & 58.5 & 10.2 & 5.4 & 15.9 & 26.1 & 23.2 & 47.0\\
OpenCLIP \cite{DBLP:conf/cvpr/ChertiBWWIGSSJ23} &  & -- & 67.0 & 71.4 & 83.6 & 89.2& 69.2& 76.1 & 50.2 & 5.7 & 8.0 & 12.4 & 25.8 & 20.4 &48.3\\
\midrule
\multicolumn{16}{c}{Tuning on 1M dataset} \\
\midrule
OpenCLIP \cite{DBLP:conf/cvpr/ChertiBWWIGSSJ23} & &  & 30.7 & 47.6 & 27.5 & 45.8  & \textbf{36.3}& 37.6 & 79.8 & 56.3 & 20.3 & 34.9 & 21.4 & 42.5  & 40.1\\
  DALIP (Ours) &\multirow{-2}{*}{ViT-B/16}  & \multirow{-2}{*}{PlantMix-1.3M}  & \textbf{36.8} & \textbf{63.0} & \textbf{48.5} & \textbf{70.5} & 34.1 & \textbf{50.6} & \textbf{85.0} & \textbf{61.3} & \textbf{33.0} & \textbf{43.8} & \textbf{23.5} & \textbf{49.3}  & \textbf{50.0}\\
\midrule
SigLIP \cite{zhai2023sigmoid} & \multirow{3}{*}{SigLIP-B/16}& \multirow{3}{*}{PlantMix-1.3M }& 51.2 & 60.0 & 71.3 & 69.9 & 52.8& 61.0&89.9 & 64.9 & 34.4 & 46.0 & 36.4 & 54.3  & 57.7\\
ProLIP* \cite{chun2024probabilistic} &  &  & 50.5 & \textbf{68.3} & 69.1 & 64.6 & \textbf{55.0} & 61.5 &  \textbf{91.3} & 64.5 & 29.6 & \textbf{49.8} & 34.0 & 53.8&57.7\\
 DALIP (Ours) &  & &   \textbf{57.2} &64.5  &\textbf{77.9} & \textbf{79.1} & 54.2 & \textbf{66.6}&90.0 & \textbf{66.9} &  \textbf{44.5} & 49.5 & \textbf{39.8} & \textbf{58.1 } & \textbf{62.4}\\
\midrule
\multicolumn{16}{c}{Tuning on 10M dataset} \\
\midrule
BioCLIP \cite{stevens2024bioclip} & \multirow{5}{*}{ViT-B/16} & TOL-10M & 18.6 & 17.9 & 0.5 & 3.9& 2.7 & 8.7&91.4& 40.7 &24.4 & 38.6 & 28.4 & 44.7  & 26.7\\
ArborCLIP-O \cite{yang2024arboretum} &  & ARBOR-40M & 11.3 & 8.0 & 0.5 & 2.4 & 1.8 &4.8 &64.5& 51.5 & 26.8 & 42.5 & 25.9 & 42.2 & 23.5\\
OpenCLIP \cite{DBLP:conf/cvpr/ChertiBWWIGSSJ23} & & PlantMix-13M & 46.8 & 61.3 & 50.2& 66.2& 50.1 &54.9 &89.9 & 47.0& 32.3 &  \textbf{48.9} & 33.0 &  50.2 & 52.6\\
FLAIR$^{\dagger}$~\cite{DBLP:journals/corr/abs-2412-03561} & & PlantMix-13M& 41.4 & 45.9 & 30.1 & 52.6 & 46.3 & 43.3 & \textbf{93.3} & 43.5 & 23.2 & 44.0 & 30.1 & 46.8 & 45.1 \\
DALIP (Ours) & &PlantMix-13M & \textbf{49.2}& \textbf{69.2} & \textbf{58.9}&  \textbf{75.2} & \textbf{55.6}& \textbf{61.6} &91.0&  \textbf{52.8}&  \textbf{34.5}&  43.7&  \textbf{34.3}& \textbf{51.3} & \textbf{56.4}\\
\midrule
OpenCLIP \cite{DBLP:conf/cvpr/ChertiBWWIGSSJ23} & &  & 55.2&  69.8 &  61.4 & 76.4  & 57.3 & 64.0& 90.4& 47.1 & 27.8 & \textbf{47.3} & 38.0&50.1&57.1 \\
DALIP (Ours) & \multirow{-2}{*}{ViT-L/14}&\multirow{-2}{*}{PlantMix-13M} & \textbf{60.6} & \textbf{72.6} & \textbf{64.0} & \textbf{83.3} & \textbf{60.6} &  \textbf{68.2} & \textbf{92.4}& \textbf{55.9}& \textbf{29.9} & 47.2 & \textbf{38.8} &\textbf{52.8} & \textbf{60.5}\\
\midrule
SigLIP~\cite{zhai2023sigmoid} & \multirow{3}{*}{SigLIP-B/16} & \multirow{3}{*}{PlantMix-13M} & 56.1 & 68.9 & 74.6 & 76.5 & 58.5 &66.9 & 95.9 & 60.6 & 40.3 & \textbf{53.5} & 46.5 & 59.4&63.1 \\
ProLIP*~\cite{chun2024probabilistic} &&  & 52.2 & \textbf{70.0} & 66.3& 67.4 & \textbf{58.9} & 63.0 &  95.0 & 55.3 & 37.5 & 44.1 & 40.1 & 54.4& 58.7\\
 DALIP (Ours) & &  &\textbf{57.1} & 69.5 & \textbf{75.4} & \textbf{78.8} & 58.6 & \textbf{67.9} & \textbf{96.5} &  \textbf{61.1} & \textbf{45.1} & 51.6 & \textbf{49.4} & \textbf{60.7} & \textbf{64.3}\\
\bottomrule
\end{tabular}
}
\caption{Zero-shot classification performance across general and plant domains for various models. ProLIP* indicates re-implementation due to the complete training code is not publicly available. FLAIR$^{\dagger}$ indicates our re-implementation by setting sub-captions to 1.}
\label{tab:Plantmix}
\end{table*}

\subsection{Implementation Details}
Following previous works~\cite{stevens2024bioclip,yang2024arboretum}, we train our DALIP models with initialization of OpenCLIP weights~\cite{radford2021learning}, where a ViT-B/16~\cite{sharir2021image} model and a 77-token causal autoregressive transformer are used as the default image encoder and text encoder, respectively. On TOL-1M and TOL-10M~\cite{stevens2024bioclip}, all models are trained within 100 epochs. For our PlantMix-13M, we train all models within 40 epochs. All models are optimized by a cosine learning rate schedule~\cite{loshchilov2016sgdr} with a batch size of 32,768. Particularly, we collect a PlantMix-1.3M to conduct ablation studies, which are randomly sampled from our PlantMix-13M. On PlantMix-1.3M, global batch size is set to 16,384. All programs are run on a server with 8 NVIDIA A100 (80GB) GPUs, and source code and models will be publicly available. More detailed settings of hyperparameters can refer to supplementary materials. 

\subsection{Results of Tuning on TREEOFLIFE}
To verify the effectiveness of our method, we first tune DALIP models on TOL-1M and TOL-10M~\cite{stevens2024bioclip}, and compare with two state-of-the-art CLIP models in biological domain, \ie, BioCLIP~\cite{stevens2024bioclip} and ArborCLIP-O~\cite{yang2024arboretum}. Specifically, our DALIP models are tuned on TOL-1M and TOL-10M, while reporting zero-shot classification performance in ten downstream biological tasks. These downstream tasks cover both animal domain (\eg, Birds525~\cite{birds525}, Plankton~\cite{sosik2015annotated}, Insects~\cite{serret2019data}, Insects 2~\cite{DBLP:conf/cvpr/WuZLCY19}, and rare species (BCR)~\cite{stevens2024bioclip}) and plant domain (\eg, PlantNet~\cite{DBLP:conf/nips/GarcinJBALCSLS21}, Fungi~\cite{DBLP:conf/wacv/PicekSMJHLF22}, PlantVillage~\cite{DBLP:journals/cee/RJ19}, Medicinal Leaf (Med. Leaf)~\cite{roopashree2020medicinal}) and PlantDoc~\cite{DBLP:conf/comad/SinghJJKK020}), aiming to evaluate zero-shot generalization ability of different models comprehensively. As shown in Table~\ref{tab:biological-domain}, our DALIP clearly outperforms BioCLIP on all downstream tasks except Birds525, PlantNet, and PlantDoc. For Birds525 and PlantNet, DALIP is comparable or superior to BioCLIP with tuning on TOL-10M, which may indicate sufficient training data is important for the two tasks. On the contrary, PlantDoc seems to be insensitive to training data and models, where all methods share comparable performance. Particularly, DALIP achieves average improvements of 3.8\% and 7.6\% over BioCLIP when tuned on TOL-1M and TOL-10M, respectively. Compared to ArborCLIP-O~\cite{yang2024arboretum} tuned on a 40M dataset, our DALIP tuned on TOL-10M achieves an improvement of 11.9\% on average of ten tasks. These results above clearly demonstrate the effectiveness of our DALIP in coping with biological data, which owes to the core idea of DALIP on optimizing CLIP models via feature distribution alignment instead of simply matching [\textit{cls}] token.

\subsection{Results of Tuning on PlantMix-13M}
Furthermore, we tune DALIP models on our PlantMix-1.3M and PlantMix-13M datasets, while comparing with several state-of-the-art CLIP models, including OpenCLIP~\cite{radford2021learning}, SigLIP~\cite{zhai2023sigmoid}, ProLIP~\cite{chun2024probabilistic} and FLAIR~\cite{DBLP:journals/corr/abs-2412-03561}. To assess effect of PlantMix dataset, all tuned models are adopted to downstream tasks in both general domain (\ie, ImageNet-1K~\cite{DBLP:conf/cvpr/DengDSLL009}, Cifar-100~\cite{krizhevsky2009learning}, Standford Cars (Cars)~\cite{gebru2017fine}, Oxford Pets (Pets)~\cite{DBLP:conf/cvpr/ParkhiVZJ12} and Sun397~\cite{DBLP:conf/cvpr/XiaoHEOT10}) and plant domain (\eg, PlantNet~\cite{DBLP:conf/nips/GarcinJBALCSLS21}, Fungi~\cite{DBLP:conf/wacv/PicekSMJHLF22}, PlantVillage~\cite{DBLP:journals/cee/RJ19}, Medicinal Leaf (Med. Leaf)~\cite{roopashree2020medicinal}) and PlantDoc~\cite{DBLP:conf/comad/SinghJJKK020}), while reporting zero-shot classification accuracy. As shown in Table~\ref{tab:Plantmix}, DALIP respectively outperforms OpenCLIP by 13.0\% and 7.2\% on average in general and plant domains when the models are tuned on PlantMix-1.3M. For SigLIP as a basic model, our DALIP-SigLIP is superior to both SigLIP and ProLIP by 4.7\% on average. 

For the models tuned on PlantMix-13M, our DALIP significantly outperforms BioCLIP \cite{stevens2024bioclip} and ArborCLIP-O \cite{yang2024arboretum} in general domain, while achieving average gains of 6.6\% and 9.1\% in plant domain, respectively. Although DALIP is inferior to the original OpenCLIP model in general domain, it achieves better average performance on general and plant domains. These results indicate that PlantMix-13M can help CLIP models achieve a better performance trade-off between plant and general domains. Comparing with the recently proposed fine-grained CLIP (\ie, FLAIR~\cite{DBLP:journals/corr/abs-2412-03561}), DALIP achieves 18.3\% and 4.5\% gains in general domain and plant-specific tasks, respectively. In addition, we compare with OpenCLIP with a larger image encoder (\ie, ViT-L), where our DALIP achieves average performance gains of 2.7\% and 4.2\% in plant and general domains. Combining with larger architecture in SigLIP, our DALIP-SigLIP performs better than the original SigLIP and the recently proposed ProLIP over 1.2\% and 5.6\% on average, respectively. These results demonstrate the effectiveness of both our PlantMix-13M dataset and DALIP method.

\subsection{Generalization to Other Specific Domains}

\begin{table}[!t]
\centering
\small
\begin{tabular*}{\columnwidth}{@{\extracolsep{\fill}}lccc@{}}
\toprule
\textbf{Model} & \textbf{AID} & \textbf{RESISC45} & \textbf{EuroSAT}\\
\midrule
GeoRSCLIP~\cite{DBLP:journals/tgrs/ZhangZGY24} & 73.7& 71.9 & 61.5 \\
DALIP (Ours)  & \textbf{75.6} & \textbf{72.9} & \textbf{64.7} \\
\midrule
\textbf{Model} & \textbf{CheXpert} & \textbf{RSNA} & \textbf{COVIDx} \\
\midrule
MGCA~\cite{DBLP:conf/nips/WangZWVY22} & 88.8 & 89.1 & 74.8 \\
DALIP (Ours) & \textbf{89.3} & \textbf{89.6} & \textbf{76.7} \\
\bottomrule
\end{tabular*}
\vspace{-0.2cm}
\caption{Results on remote sensing and medical imaging domains.}
\vspace{-0.3cm}
\label{tab:cross_domain}
\end{table}

To further verify the generalization ability of our DALIP, we conduct experiments on remote sensing and medical imaging domains. For remote sensing domain, following GeoRSCLIP~\cite{DBLP:journals/tgrs/ZhangZGY24}, we fine-tune DALIP on RS5M and evaluate zero-shot classification on AID~\cite{DBLP:journals/tgrs/XiaHHSBZZL17}, RESISC45~\cite{DBLP:journals/pieee/ChengHL17}, and EuroSAT~\cite{DBLP:journals/staeors/HelberBDB19}. As shown in Table~\ref{tab:cross_domain}, DALIP achieves 75.6\%, 72.9\%, and 64.7\% accuracy, outperforming GeoRSCLIP by 1.9\%, 1.0\%, and 3.2\%. In medical imaging domain, we pre-train DALIP on MIMIC-CXR 2.0~\cite{DBLP:journals/corr/abs-1901-07042} following the settings of MGCA~\cite{DBLP:conf/nips/WangZWVY22}, where BioClinicalBERT~\cite{alsentzer2019publicly} and an MBDC-enhanced ViT-B-16 are combined as the basic CLIP model. Then, a linear classifier in~\cite{DBLP:conf/nips/WangZWVY22} learned on 1\% training data is used to evaluate performance  on downstream tasks. As shown in Table~\ref{tab:cross_domain}, DALIP achieves 89.3\%, 89.6\% and 76.7\% AUROC on CheXpert~\cite{irvin2019chexpert}, RSNA~\cite{shih2019augmenting} and COVIDx~\cite{DBLP:journals/corr/abs-2003-09871}, surpassing MGCA~\cite{DBLP:conf/nips/WangZWVY22} by 0.5\%, 0.5\%, and 1.9\%. They show the effectiveness of our DALIP on various specific domains.

\subsection{Ablation Study}\label{sec:ablationStudy}
In this subsection, we conduct ablation studies to assess the effect of first (1st)- and second (2nd)-order statistics on DALIP, while comparing our Multi-head Brownian Distance Covariance (MBDC) with several counterparts on PlantMix-1.3M dataset. Finally, we evaluate the effect of data mixing ratio on PlantMix-10M dataset. Specifically, we first tune the models on PlantMix datasets, and then evaluate on general domain (\ie, IN-1K~\cite{DBLP:conf/cvpr/DengDSLL009}) and plant domain (average of five downstream datasets, including PlantNet~\cite{DBLP:conf/nips/GarcinJBALCSLS21}, Fungi~\cite{DBLP:conf/wacv/PicekSMJHLF22}, PlantVillage~\cite{DBLP:journals/cee/RJ19}, Med. Leaf~\cite{roopashree2020medicinal}) and PlantDoc~\cite{DBLP:conf/comad/SinghJJKK020}).

\begin{table}
\centering
\small
\begin{tabular*}{\columnwidth}{@{\extracolsep{\fill}}lccc@{}}
\toprule
\textbf{Model} &\textbf{Dataset}&
\textbf{IN-1K}
& \textbf{Plant Mean}
\\
\midrule
\multicolumn{4}{c}{Effect of 1st-\&2nd-Order Statistics} \\
\midrule
 DALIP$_{1st}$ &\multirow{2}{*}{PlantMix-1.3M} & 30.7& 42.5\\
DALIP$_{2nd}$ & & \textbf{32.9}  & \textbf{44.8} \\

\midrule

\multicolumn{4}{c}{Comparison of Second-Order Counterparts} \\
\midrule

 DALIP$_{\text{MP}}$& \multirow{4}{*}{PlantMix-1.3M}&  36.0 & 43.6 \\
DALIP$_{\text{BDC}}$& & 36.1  & 46.9  \\
DALIP$_{\text{KSPD}}$ & & 31.9 & 42.8 \\
{DALIP$_{\text{MBDC}}$} & &  \textbf{36.8} & \textbf{49.3}  \\
\midrule
\multicolumn{4}{c}{Effect of Data Mixing} \\
\midrule
\multirow{5}{*}{DALIP} &P: 10M +  G: 0M & 18.6  & 52.0  \\
 &P: 10M +  G: 1M & 43.1 & 51.9 \\
 &P: 10M +  G: 2M & 47.8 & 51.6  \\
 &P: 10M +  G: 3M & 49.2 &51.3  \\
 &P: 10M +  G: 4M  & 49.3 & 48.1 \\
\bottomrule
\end{tabular*}
\vspace{-0.2cm}
\caption{Ablation Study for DALIP and PlantMix. ``P: 10M + G: 0M" indicates that the dataset consists of 10 million plant data  without general data.}
\vspace{-0.3cm}
\label{tab:ablation}
\end{table}
\noindent\textbf{Effect of 1st-\&2nd-order Statistics.}  As shown in the top part of Table~\ref{tab:ablation}, DALIP with only 2nd-order statistics (\ie, DALIP$_{2nd}$) outperforms one with only 1st-order statistics (\ie, DALIP$_{1st}$) by 2.2\% and 2.3\% on general and plant domains, respectively. Our DALIP with both 1st-\&2nd-order statistics boosts performance to 36.8\% and 49.3\% in general and plant domains, indicating strong complementarity between 1st- and 2nd-order statistics.

\noindent\textbf{Comparison of Second-order Counterparts.}
To assess the effect of our MBDC for realizing second-order representations, we compare with two covariance counterparts (\ie, Moment Probing (MP) \cite{DBLP:conf/iccv/GaoWLZHZ23} and Brownian Distance Covariance (BDC) \cite{DBLP:conf/cvpr/XieLLWL22}) and a kernel method (KSPD) \cite{engin2018deepkspd} by replacing our MBDC in DALIP with similar representation sizes.  As shown in the middle of Table~\ref{tab:ablation}, our DALIP with MBDC respectively outperforms MP (DALIP$_\text{MP}$) , BDC (DALIP$_\text{BDC}$) and KSPD (DALIP$_\text{DeepKSPD}$) by 0.8\% , 0.7\% and 4.9\% on IN-1K, while being superior to them over 5.7\% , 2.4\% and 6.5\% in plant domain. These results demonstrate the effectiveness of our proposed MBDC.

\noindent\textbf{Effect of Data Mixing.}
To further assess the effect of data mixing on PlantMix-13M dataset, we conduct experiments by gradually increasing the general domain data (from 0M to 4M) on 10M plant data. As shown in the bottom of Table~\ref{tab:ablation}, general domain performance gradually improves as general-domain data increases, reaching 49.2\% and 49.3\% with 3M and 4M added general data. Conversely, performance in plant domain  steadily declines, dropping from 51.3\% to 48.1\% when 4M general data are added. It is consistent with our observations in Sec. \ref{sec:plantmix}, and further supports our conclusion in data mixing laws. 
\section{Conclusion}
Concerning domain-specific CLIP models for biological data, this paper proposed a Distribution Alignment-based Language-Image Pre-Training (DALIP) method to effectively cope with fine-grained nature of biological data. Specifically, DALIP models feature distribution of image-text pairs as powerful representations, and optimizes CLIP models by aligning distribution of image-text pairs. For computational efficiency, DALIP approximates feature distribution via its first- and second-order statistics, while a MBDC module is presented to acquire second-order statistics. Furthermore, a PlantMix-13M is collected for plant domain according to data mixing laws. Extensive experiments clearly demonstrate the effectiveness of our DALIP in biological domain, while our PlantMix-13M dataset further helps DALIP achieve better trade-offs in plant and general domains. In the future, we will adopt our DALIP to other domain-specific data~\cite{DBLP:conf/cikm/ShinPW0OS22}, while investigating to integrate regularization techniques for further alleviating the issue of catastrophic forgetting.

\renewcommand\thesection{S\arabic{section}}
\setcounter{section}{0}
\setcounter{equation}{0}
\renewcommand\theequation{S.\arabic{equation}}
\renewcommand\thefigure{S\arabic{figure}}
\setcounter{figure}{0}
\renewcommand\thetable{S\arabic{table}}
\setcounter{table}{0}

\section{Complexity Analysis}
\begin{figure}[ht!]
\centering
\includegraphics[width=0.42\textwidth]{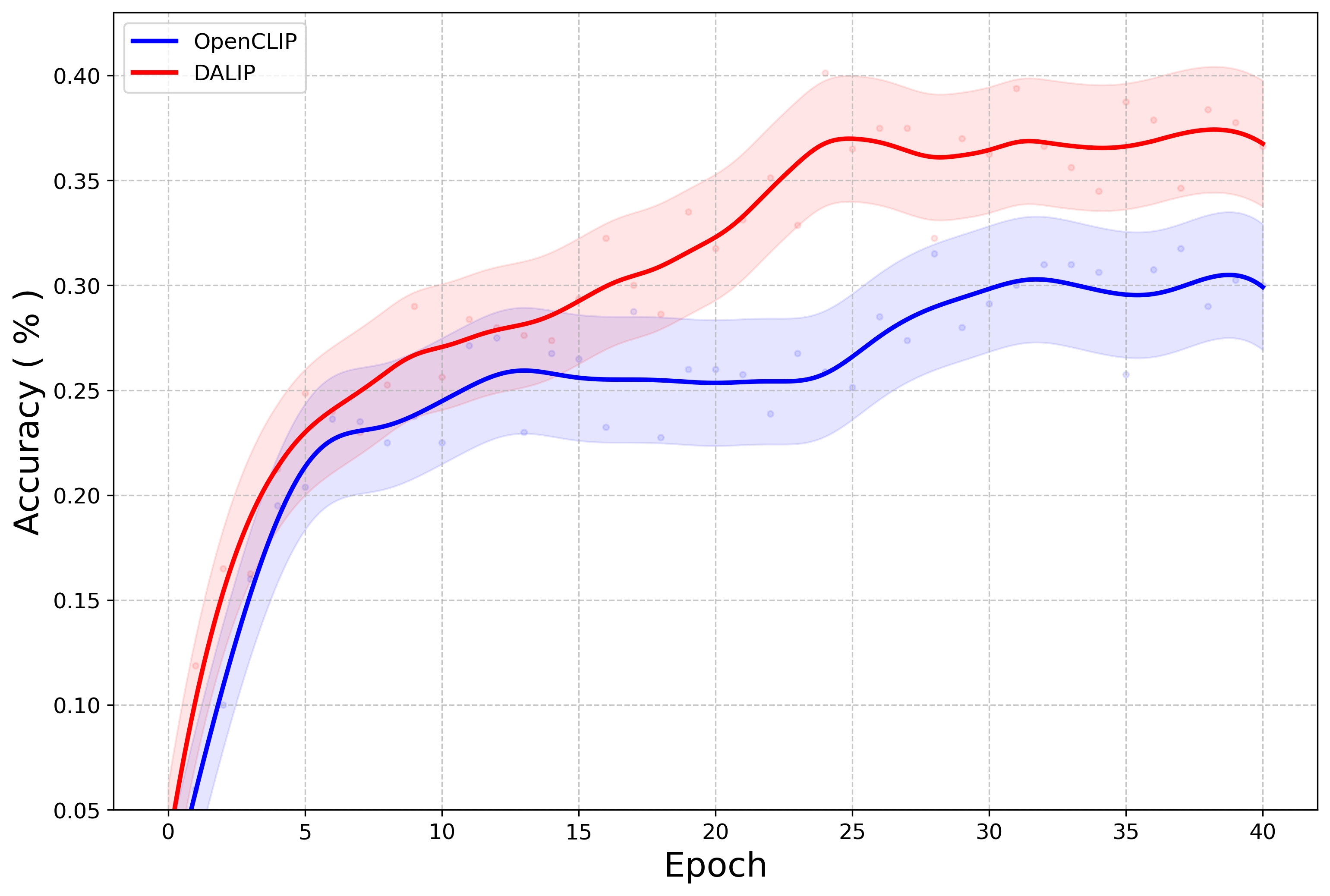}
  \caption{Convergence speed for DALIP and OpenCLIP with tuning on TOL-1M, where accuracies on Fungi are reported. For briefness, we show the results within the first 40 training epochs.}
  \label{fig:111}

\end{figure}

To analyze the computational efficiency of our DALIP, we compare it with the original CLIP~\cite{radford2021learning} in terms of training time per batch, inference time per image, and convergence speed. All comparisons are conducted on 8 NVIDIA A100 GPUs with batch sizes of 2048 and 256 for training and testing, respectively. As shown in  Table~\ref{table:speed_comparison}, our DALIP increases training time by 0.03 seconds per batch over the original CLIP. However, as shown in Figure~\ref{fig:111}, DALIP has a faster convergence speed than the original CLIP. Notably, DALIP tuned within about 20 epochs achieves comparable results  with the original CLIP tuned within 40 epochs, which helps the models reach an expected result by using fewer training epochs and reduces training time. For inference, DALIP brings an additional 0.93 ms per image, which is affordable for practical applications. In conclusion, DALIP can achieve a better trade-off between efficiency and effectiveness.

\begin{table}[ht!]
\centering
\setlength{\tabcolsep}{2pt} 
\small 
\begin{tabular}{lccc}
\toprule
Models & Qwen2-VL-7B & Qwen2-VL-72B & InternVL2-8B \\
\midrule
Acc. (\%) & 91 & 77 & 87 \\
Latency (s) & 0.1 & 1.0 & 0.2 \\
\bottomrule
\end{tabular}
\caption{Comparison of Caption Generation Quality and Generation Time. The captions are generated using various open-source VLLMs and evaluated by GPT-4o to assess their accuracy and identify potential hallucinations. Latency is measured as the average time required to produce a single caption.
}
\label{tab:VLM-generatedQuality}
\end{table}

\section{More Examples on Generated Descriptions}
\begin{figure*}[ht!]
\centering
\includegraphics[width=0.84\textwidth]{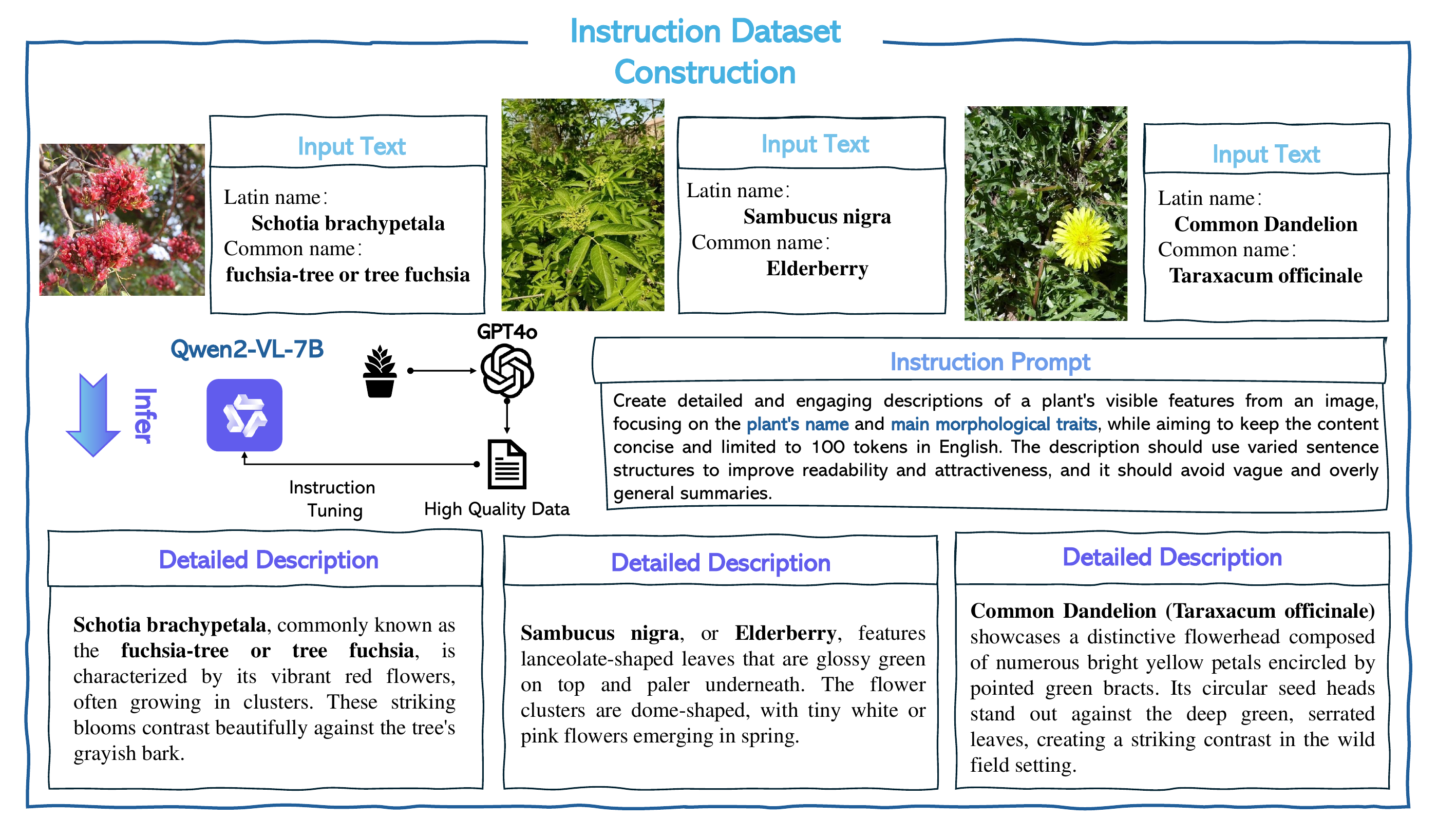}
  \caption{More examples of generating precise and detailed plant descriptions using qwen2VL-7B, based on Latin and Common names,
images, and tailored instruction prompts.}
  \label{fig:datasetsamples}
\end{figure*}

As illustrated in Fig.~\ref{fig:datasetsamples}, we show more examples on the generation of precise and detailed plant  descriptions by prompting Qwen2VL-7B with Latin and common names, images, and customized instructions. Clearly, generated descriptions are different from those utilized by CLIP.

\section{Quality of MLLM-generated Captions}
To ensure the quality of generated captions while maintaining computational efficiency, we conducted a systematic evaluation of open-source MLLMs. We compare caption quality and inference efficiency of Qwen2-VL-7B \cite{Qwen2VL}, Qwen2-VL-72B \cite{Qwen2VL} and InternVL2-8B \cite{chen2024far}, aiming to balance practicality and scalability under budget constraints. While GPT-4o achieves SOTA performance, its cost is prohibitive for large-scale datasets: our PlantMix with 13M images would require approximately \$150k. To validate caption accuracy and detect hallucinations, we sampled 1K generated captions (on 8 NVIDIA A100 GPUs) and evaluated them using GPT-4o. As shown in Tab \ref{tab:VLM-generatedQuality}, Qwen2-VL-7B achieves 91\% accuracy and clearly surpasses Qwen2-VL-72B (77\%) and InternVL2-8B (87\%), Qwen2-vl-7B generates a caption in an average of 0.1s, significantly faster than Qwen2-vl-72B, which takes 1.0s per caption. It also outperforms InternVL2-8B by 0.1s in generation speed.

\begin{table}[ht!]
\centering
\setlength{\tabcolsep}{2pt} 
\renewcommand{\arraystretch}{0.8}
\begin{tabular}{lcc}
\toprule
 & Training Time (s) & Inference Time (ms) \\
\midrule
CLIP & 1.64 &  4.02\\
DALIP (Ours)& 1.67 & 4.95 \\
\bottomrule
\end{tabular}
\caption{Computational complexity of DALIP
and CLIP in terms of training time per batch (s) and inference time per image (ms).}
\label{table:speed_comparison}
\end{table}

\section{Detailed Results of Ablation Study}
As shown in Table~\ref{tab:ablationdetail}, we give the detailed results on five plant domain tasks (\eg, PlantNet, Fungi, PlantVillage, Medicinal Leaf (Med. Leaf) and PlantDoc, as discussed in Sec. 5.4 of manuscript. From it, we can see that performance on each dataset is consistent with the average.

\section{More Ablation Studies}
Here, we further assess the effect of input resolution and visual encoder on our DALIP.

\noindent\textbf{Comparison of Larger Resolution.}
As shown in the middle of Table~\ref{tab:ablationdetail}, we compare CLIP with DALIP models by using two input resolutions. Specifically, we take the models tuned on 224x224 resolution inputs and tune them for 10 epochs using 336x336 resolutions continuously. The results demonstrate that increasing the resolution improves performance for both CLIP and DALIP. CLIP$_{336}$ achieves a 0.7\% increase in mean performance over CLIP$_{224}$, while DALIP$_{336}$ shows a more substantial 1.4\% improvement over DALIP$_{224}$. Notably, DALIP$_{336}$ outperforms CLIP$_{336}$ by 2.5\% on average (51.0\% vs. 49.2\%).

\begin{table*}[ht!]
\centering
\small
\renewcommand{\arraystretch}{0.6} 
\begin{tabular}{lccccccccccc}
\toprule
\multirow{10}{*}{\textbf{Model}} & \multirow{10}{*}{\textbf{Dataset}} &  & \multicolumn{5}{c}{\textbf{Plants \& Fungi}}& \multirow{10}{*}{Plant Mean}& \multirow{10}{*}{Mean}\\
\cmidrule(lr){4-8}
 & & \rotatebox[origin=c]{90}{Imagenet} & \rotatebox[origin=c]{90}{PlantNet} & \rotatebox[origin=c]{90}{Fungi} & \rotatebox[origin=c]{90}{PlantVillage} & \rotatebox[origin=c]{90}{Med. Leaf} & \rotatebox[origin=c]{90}{PlantDoc} & \\
\midrule
\multicolumn{10}{c}{Effect of 1st-\&2nd-Order Statistics} \\
\midrule
DALIP$_{1st}$ & \multirow{2}{*}{PlantMix-1.3M}& 30.7 & 79.8 & 56.3 & 20.3 & 34.9 &  21.4 & 42.5&36.6 \\
DALIP$_{2nd}$ & & \textbf{32.9}  & \textbf{81.1} & \textbf{58.4} & \textbf{24.2} & \textbf{36.8} & \textbf{23.5} & \textbf{44.8} & \textbf{38.9} \\

\midrule
\multicolumn{10}{c}{Comparison of Second-Order Counterparts} \\
\midrule
DALIP$_{\text{MP}}$ & \multirow{4}{*}{PlantMix-1.3M}& 36.0 & 78.4 & 54.3 & 28.7 & 38.2 & 18.4 & 43.6&39.8 \\
DALIP$_{\text{BDC}}$ & & 36.1  & 82.4 & 58.0 & 31.5 & 41.5 & 21.6 & 46.9&41.5 \\
DALIP$_{\text{DeepKSPD}}$ & & 31.9  &  78.8 & 55.2 & 24.7 & 35.1 & 20.2 & 42.8 & 37.4 \\
DALIP & & \textbf{36.8} & \textbf{85.0} & \textbf{61.3} &  \textbf{33.0} & \textbf{43.8} & \textbf{23.5} & \textbf{49.3} & \textbf{43.1} \\
\midrule
\multicolumn{10}{c}{Comparison of Larger Resolution} \\
\midrule
CLIP$_{224}$ & \multirow{4}{*}{PlantMix-13M} & 46.8 & 89.9 & 47.0& 32.3 &  48.9 & 33.0 &  50.2 &  48.5\\
CLIP$_{336}$ & & 47.2 & 90.8 &  48.3 & 33.6 & \textbf{49.2} & 33.5& 51.1 & 49.2 \\
DALIP$_{224}$& & 49.2 & 91.0 & 52.8 & 34.5 & 43.7 & \textbf{34.3} & 51.3&50.3\\
DALIP$_{336}$ & & \textbf{50.1} & \textbf{91.6} & \textbf{53.5} & \textbf{35.4} & 44.6 & 34.2 & \textbf{51.9} & \textbf{51.0} \\
\midrule
\multicolumn{10}{c}{Backbone of ConvNEXT-base} \\
\midrule
CLIP$_{\text{ConvNEXT}}$ & \multirow{2}{*}{PlantMix-13M} & 46.5 & 89.6 & 45.3 & 30.1 & \textbf{47.8} & 32.6 &  49.1 &  47.8\\
DALIP$_{\text{ConvNEXT}}$ &  & \textbf{48.2} & \textbf{90.2} & \textbf{47.0} & \textbf{32.6}  & 47.0 & \textbf{33.7} & \textbf{50.1} & \textbf{49.2} \\
\midrule
\multicolumn{10}{c}{Effect of Data Mixing} \\
\midrule
\multirow{5}{*}{DALIP} & P: 10M +  G: 0M & 18.6  & \textbf{93.0} & \textbf{53.7} & \textbf{36.4} & 40.0 & \textbf{36.9} & \textbf{52.0} &35.3\\
 & P: 10M +  G: 1M & 43.1 & 91.8 & 53.0 & 34.5 & \textbf{45.6} & 34.6 & 51.9& 47.5\\
 &P: 10M +  G: 2M & 47.8 & 91.6 & 53.2 & 34.8 & 44.0 & 34.4 & 51.6& 49.7\\
 &P: 10M +  G: 3M& 49.2 & 91.0 & 52.8 & 34.5 & 43.7 & 34.3 & 51.3& \textbf{50.3} \\
 &P: 10M +  G: 4M & \textbf{49.3} & 90.1 & 48.6 & 30.8 & 39.3 & 31.7 & 48.1& 48.7\\
\bottomrule
\end{tabular}
\caption{Ablation Study of PlantMix and DALIP}
\label{tab:ablationdetail}
\end{table*}

\begin{figure*}[ht!]
\vspace{-0.3cm}
\centering
\includegraphics[width=0.99\textwidth]{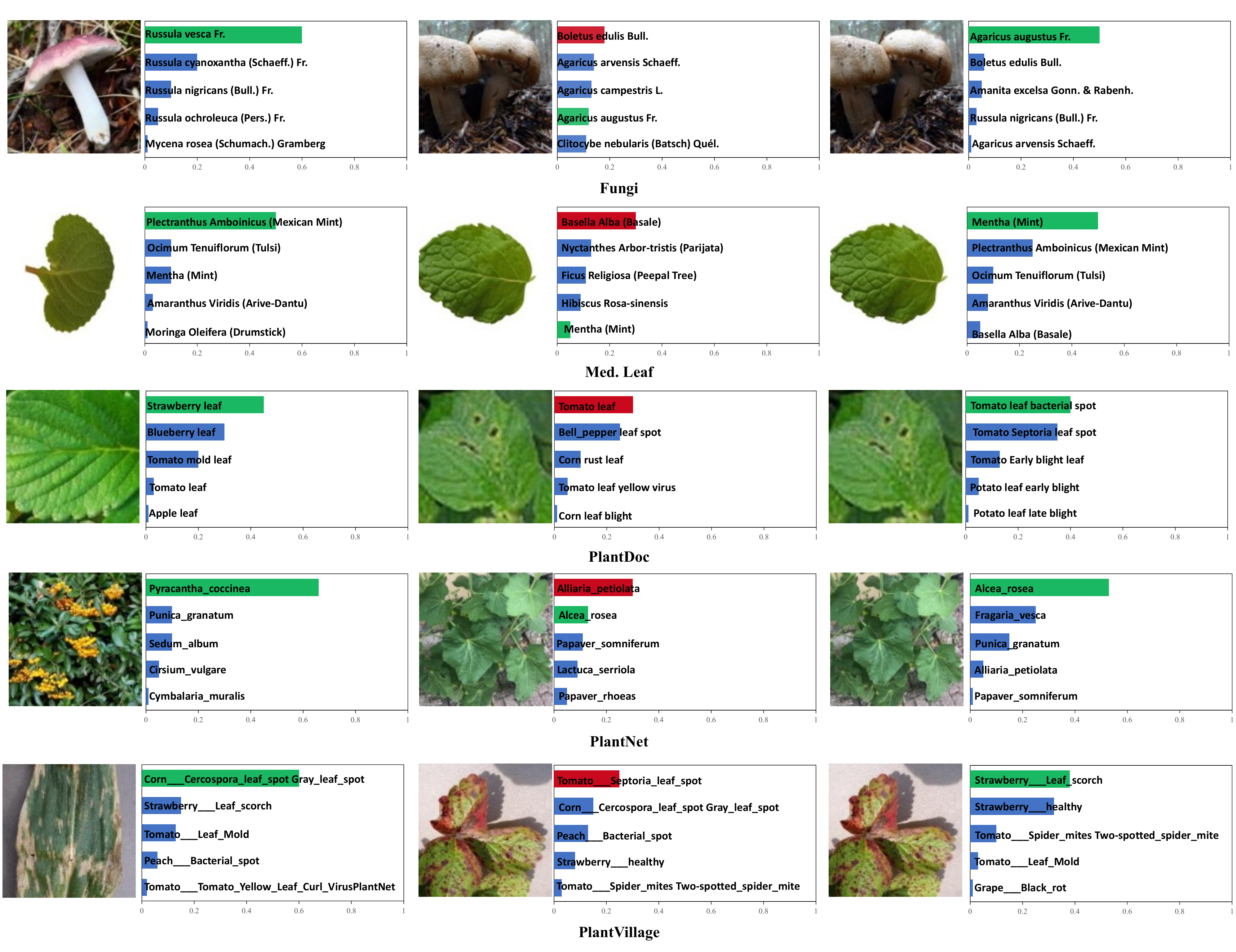}
\vspace{-0.3cm}
\caption{Example predictions for DALIP and BioCLIP on Fungi, Med. Leaf, PlantDoc, PlantNet and PlantVillage. Ground truth
labels are green; incorrect predictions are red. Left: Correct DALIP predictions. Center, Right: Images that BioCLIP incorrectly labels, but DALIP correctly labels.}
  \label{fig:predictions}
\end{figure*}

\begin{table*}[h]
\centering
\begin{tabular}{ccccccccc}
\toprule
$\lambda_1$ & 0.0 & 0.2 & 0.3 & 0.4& 0.5 & 0.6 & 0.8 & 1.0\\
\midrule
Acc (\%) & 44.8 & 47.5 & 48.8 & 49.3 & 48.6 & 48.1 & 46.8 & 42.5 \\
\bottomrule
\end{tabular}
\caption{Results of various $\lambda$ on Plant (Mean). $\lambda_1 + \lambda_2 = 1$.}
\label{tab:lambda_sensitivity_horizontal}
\end{table*}

\noindent\textbf{Visual Encoder of ConvNEXT-base.}
To further assess the effect of visual encoder, we employ ConvNEXT-base as an alternative to the ViT-B/16 backbone, and compare  CLIP with DALIP. As shown in the last two rows of Table~\ref{tab:ablationdetail}, DALIP with ConvNEXT-base backbone outperforms CLIP$_{\text{ConvNEXT}}$  across all plant domain tasks. The average performance of DALIP$_{\text{ConvNEXT}}$ is 1.4\% higher than that of CLIP$_{\text{ConvNEXT}}$, demonstrating the effectiveness of our DALIP across  different visual encoders. Notably, DALIP$_{\text{ConvNEXT}}$ shows significant improvements in tasks such as Fungi and PlantDoc, leading by 1.7\% and 1.1\% respectively. These results suggest that DALIP is not limited to a specific backbone architecture but can be generalized to other architectures  like ConvNEXT.

\noindent\textbf{Sensitivity analysis for $\lambda_{1}$ and $\lambda_{2}$.}
 To investigate how varying $\lambda_{1}$ (with $\lambda_{1}+\lambda_{2}$ = 1) affects model performance on Plant test set, we conduct experiments using multiple different values of $\lambda_{1}$. As shown in Table~\ref{tab:lambda_sensitivity_horizontal}, performance remains stable for $\lambda_{1}$ values between 0.3 and 0.6 (48.1\%–49.3\%), demonstrating robustness to parameter variations in this range. However, the 8.8\% performance gap between optimal ($\lambda_{1}$=0.4) and worst-case ($\lambda_{1}$=1.0) configurations underscores the necessity of avoiding unbalanced weightings.


\section{Comparison of Prediction Examples}
Fig.~\ref{fig:predictions} presents a comparative analysis of zero-shot prediction examples between DALIP and BioCLIP across five plant domain tasks, wherein DALIP consistently exhibits superior performance. DALIP's enhanced feature extraction capabilities enable it to discern subtle morphological nuances in fungi, leaf structures, and disease symptoms. This refined ability is particularly evident in its accurate identification of Agaricus augustus Fr. and Alcea rosea, where BioCLIP struggles. Furthermore, DALIP demonstrates a more nuanced contextual understanding, accurately diagnosing plant diseases such as tomato leaf bacterial spot and strawberry leaf scorch. This suggests a heightened capacity to correlate visual cues with specific conditions. Notably, DALIP's consistent accuracy across diverse datasets (Fungi, Med. Leaf, PlantDoc, PlantNet, and PlantVillage) indicates the development of a more comprehensive and transferable knowledge base for plant-related tasks. These results collectively underscore DALIP's advanced proficiency in zero-shot learning scenarios within the plant domain.

\small
\bibliographystyle{ieeenat_fullname}
\bibliography{main}
\end{document}